\pdfoutput=1

\documentclass[11pt]{article}

\usepackage[final]{acl}

\usepackage{times}
\usepackage{latexsym}

\usepackage[T1]{fontenc}

\usepackage[utf8]{inputenc}

\usepackage{microtype}

\usepackage{inconsolata}

\usepackage{graphicx}

\usepackage{scrhack}
\usepackage{listings}
\usepackage{lstautogobble}
\usepackage{xcolor}  
\usepackage[utf8]{inputenc}
\usepackage{caption}

\usepackage{acronym}   
\usepackage{booktabs} 
\usepackage{tabularx} 
\usepackage[Export]{adjustbox}
\usepackage{siunitx}    
\usepackage{threeparttablex}    
\usepackage{textcomp}           
\usepackage{placeins}
\usepackage{multirow}

\definecolor{TUMBlue}{HTML}{0065BD}
\definecolor{TUMSecondaryBlue}{HTML}{005293}
\definecolor{TUMSecondaryBlue2}{HTML}{003359}
\definecolor{TUMBlack}{HTML}{000000}
\definecolor{TUMWhite}{HTML}{FFFFFF}
\definecolor{TUMDarkGray}{HTML}{333333}
\definecolor{TUMGray}{HTML}{808080}
\definecolor{TUMLightGray}{HTML}{CCCCC6}
\definecolor{TUMAccentGray}{HTML}{DAD7CB}
\definecolor{TUMAccentOrange}{HTML}{E37222}
\definecolor{TUMAccentGreen}{HTML}{A2AD00}
\definecolor{TUMAccentLightBlue}{HTML}{98C6EA}
\definecolor{TUMAccentBlue}{HTML}{64A0C8}

\newacro{LLM}{large language model}
\newacro{TOD}[ToD]{task-oriented dialogue system}
\newacro{MWOZ}[MultiWOZ]{Multi-Domain Wizard-of-Oz dataset}
\newacro{SGD}[SGD]{Schema-Guided Dialogue dataset}
\newacro{USS}[USS]{User Satisfaction Simulation dataset}
\newacro{JDDC}[JDDC]{Jing Dong Dialogue Corpus}
\newacro{DBDC}[DBDC]{Dialogue Breakdown Detection Challenge}
\newacro{DSTC}[DSTC]{Dialogue System Technology Challenge}
\newacro{NB}[NB]{not a breakdown}
\newacro{PB}[PB]{possible breakdown}
\newacro{B}[B]{breakdown}
\newacro{CLI}[CLI]{command line interface}
\newacro{CE}[CE]{conversational error type}
\newacro{TE}[TE]{task-oriented error type}
\newacro{EM}[EM]{exact match}
\newacro{SM}[SM]{superset match}
\newacro{PM}[PM]{partial match}
\newacro{MTLD}[MTLD]{measure of textual lexical diversity}
\newacro{UT}[$\#$UT]{total number of user turns}
\newacro{ST}[$\#$ST]{total number of system turns}
\newacro{STperD}[ST/D]{average number of system turns per dialogue}
\newacro{UTL}[$|UT|$]{user turn length in words}
\newacro{STL}[$|ST|$]{system turn length in words}
\newacro{DwithB}[$\#$D with B]{total number of dialogues with breakdowns}
\newacro{TotalB}[$\#$B]{total number of breakdowns}
\newacro{BperST}[B/ST]{average number of breakdowns per system turn}
\newacro{UniqueB}[$\#$Unique B]{number of unique error types}
\newacro{TurnsToB}[Avg. TtB]{average number of turns until the first breakdown}
\newacro{CCrash}[$\#$Crash]{total number of dialogue system crashes}
\newacro{TwithM}[$\#$Testers with match]{number of breakdown testers where the respective error type was triggered}
\newacro{AutoTODSim}[AutoTOD-Sim]{AutoTOD User Simulator}
\newacro{PerSim}[PerSim]{Persona Simulator}
\newacro{StdPerSim}[Std-PerSim]{Standard Persona Simulator}
\newacro{ChallPerSim}[Chall-PerSim]{Challenging Persona Simulator}

\newcommand*{\lenientBLabel}{B\textsuperscript{+}}
\newcommand*{\lenientNBLabel}{NB\textsuperscript{-}}

\newcommand*{\chatChecker}{\texttt{ChatChecker}}
\newcommand*{\breakdownDetector}{\textit{Breakdown Detector}}
\newcommand*{\dialogueRater}{\textit{Dialogue Rater}}
\newcommand*{\personaGenerator}{\textit{Persona Generator}}

\newcommand*{\gptthreeturbo}{\texttt{GPT-3.5 Turbo}}
\newcommand*{\gptfour}{\texttt{GPT-4}}
\newcommand*{\gptfourturbo}{\texttt{GPT-4 Turbo}}
\newcommand*{\gptfourO}{\texttt{GPT-4o}}

\newcommand*{\autotod}{AutoTOD}
\newcommand*{\goalSettingBot}{Goal-Setting Assistant}

\newcommand*{\ghassel}{\emph{ghassel}}

\title{ChatChecker: A Framework for Dialogue System Testing and Evaluation Through Non-cooperative User Simulation}

\author{Roman Mayr \\
  Technicial University of Munich  \\
  \& University of Cambridge \\
  \texttt{roman.mayr@tum.de} \\\And
  Michel Schimpf \\
  University of Cambridge \\
  \texttt{ms2957@cam.ac.uk} \\
  \\\And
  Thomas Bohné \\
  University of Cambridge \\
  \texttt{tbm35@cam.ac.uk} \\}

\begin{document}
\maketitle
\begin{abstract}
While modern dialogue systems heavily rely on \acp{LLM}, their implementation often goes beyond pure \ac{LLM} interaction. Developers integrate multiple \acp{LLM}, external tools, and databases. Therefore, assessment of the underlying \ac{LLM} alone does not suffice and the dialogue systems must be tested and evaluated as a whole. However, this remains a major challenge. With most previous work focusing on turn-level analysis, less attention has been paid to integrated dialogue-level quality assurance. 
To address this, we present \chatChecker{}\footnote{\url{https://github.com/llm-psychology-group/chat-checker}}, a framework for automated evaluation and testing of complex dialogue systems. \chatChecker{} uses \acp{LLM} to simulate diverse user interactions, identify dialogue breakdowns, and evaluate quality. Compared to previous approaches, our design reduces setup effort and is generalizable as it does not require reference dialogues and is decoupled from the implementation of the target dialogue system.
We improve breakdown detection performance over a prior \ac{LLM}-based approach by including an error taxonomy in the prompt.
Additionally, we propose a novel non-cooperative user simulator based on challenging personas that uncovers weaknesses in target dialogue systems more effectively.
Through this, \chatChecker{} contributes to thorough and scalable testing. This enables both researchers and practitioners to accelerate the development of robust dialogue systems.

\end{abstract}

\section{Introduction}

Dialogue-based human-computer interaction has become increasingly widespread with the rise of \acp{LLM} through systems such as ChatGPT~\cite{openai_chatgpt_2024}. Beyond the chat interfaces of the \ac{LLM} providers, dialogue systems are now deployed across various use cases - from task-oriented customer service, over mental health treatment~\cite{heinz_randomized_2025} to conversations with virtual companions like Replika~\cite{replika2025}.

With millions of users and applications in critical domains such as healthcare and finance, ensuring the robustness and reliability of dialogue systems is crucial. However, testing and evaluating these systems remain persistent challenges~\citep{yoshino_overview_2023, rodriguez-cantelar_overview_2023}.

\begin{figure}
    \centering
    \includegraphics[width=1.0\linewidth]{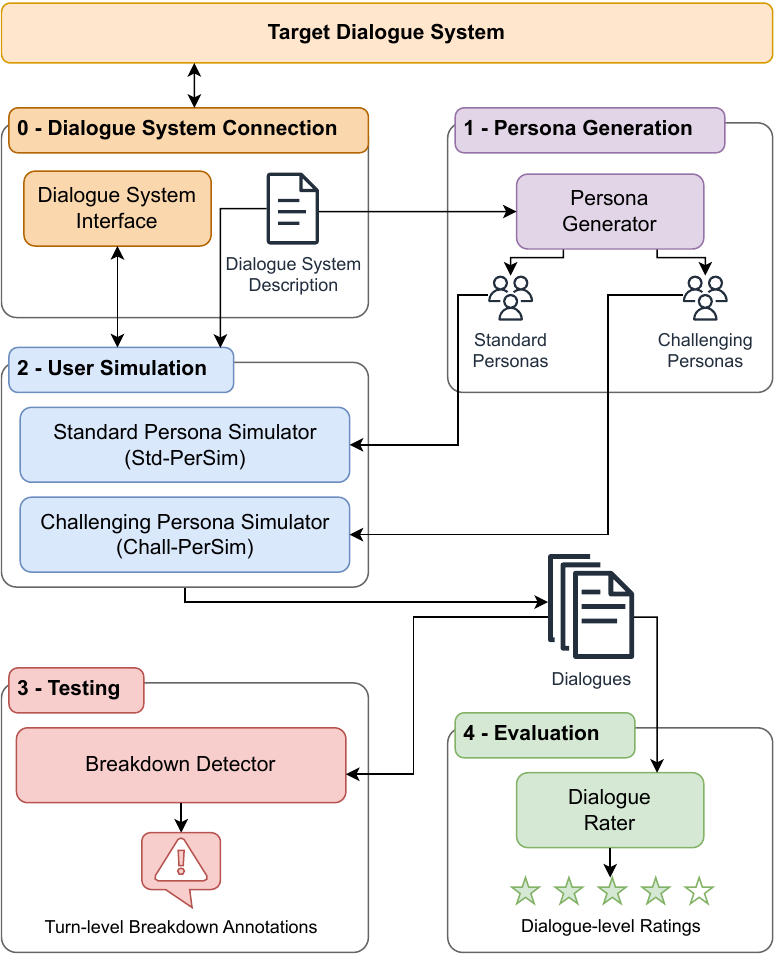}
    \caption{Schematic overview of \chatChecker{}. After connecting their target dialogue system, developers can generate user personas, run simulations, identify dialogue breakdowns, and obtain ratings.}
    \label{fig:chatchecker-overview}
\end{figure}

\subsection{Dialogue Systems}\label{subsec:dialogue-systems}
Dialogue systems, also called chatbots, interact with users through natural language conversation~\citep{jurafsky_speech_2025}. These systems can be broadly categorized into \acp{TOD} designed for specific tasks like booking hotels, and conversational dialogue systems for open-domain, natural conversations \citep{deriu_survey_2021}. Modern dialogue systems increasingly rely on \acp{LLM}~\citep{xu_rethinking_2024, yi2024surveyrecentadvancesllmbased}, which blur the traditional distinction between these categories~\citep{jurafsky_speech_2025}.

\subsection{Dialogue System Evaluation}\label{subsec:dialogue-evaluation}
Evaluating dialogue systems encompasses both human and automatic assessment of system performance. While human evaluation through crowd-sourcing platforms is common, its cost and time requirements drive the need for automated methods~\citep{deriu_survey_2021}.

Traditional reference-based metrics like BLEU \citep{papineni_bleu_2002} have proven inadequate for dialogue evaluation~\citep{liu_how_2016}. Recent research leverages \acp{LLM} for automated rating, with approaches such as G-EVAL \citep{liu_g-eval_2023} showing improved correlations with human judgments. \citet{mendonca_simple_2023} demonstrated that LLM-based ratings achieve state-of-the-art performance for multilingual dialogue evaluation.

\subsection{Dialogue Breakdown Detection}\label{subsec:breakdown-detection}
In this work, we focus on the elicitation of undesired interactions for testing dialogue systems. To this end, it is crucial to detect dialogue breakdowns, which occur when a conversation becomes difficult to continue smoothly \citep{martinovsky_error_2003, higashinaka_fatal_2015}. The \ac{DBDC}~\citep{higashinaka_dialogue_2016} provides datasets labeled as \ac{NB}, \ac{PB}, or \ac{B}. For a more fine-grained assessment, \citet{higashinaka_integrated_2021} developed a comprehensive error taxonomy distinguishing 17 conversational error types across utterance, response, context, and society levels.
\citet{ghassel_are_2024} reported that \gptfour{} achieved competitive results in breakdown detection, establishing \acp{LLM} as effective tools for this task.

\subsection{User Simulation}\label{subsec:user-simulation}
User simulators automate the generation of dialogue interactions for testing and evaluation. While early approaches like ABUS~\citep{schatzmann_agenda-based_2007} focused on semantic-level simulation, recent methods directly generate user utterances using \acp{LLM}~\citep{terragni_-context_2023, davidson_user_2023, xu_rethinking_2024}.

Although these existing \ac{LLM}-based simulators demonstrate the potential of leveraging \acp{LLM} for user simulation, they have significant limitations:
\begin{itemize}
    \item \emph{Dependence on existing datasets}: They typically rely on existing reference dialogues for few-shot samples and structured goals. Early-stage systems or new iterations often lack sufficient interaction data for this.
    \item \emph{Tight coupling with the target dialogue system}: For instance, \citet{terragni_-context_2023} integrated their simulator in the ConvLab-2 framework~\citep{zhu_convlab-2_2020} for dialogue systems on the \ac{MWOZ}~\citep{budzianowski_multiwoz_2018, eric_multiwoz_2020} benchmark.
    \item \emph{Focus on cooperative users}: With reference dialogues extracted from human samples and \acp{LLM} instructed to attempt to achieve a given in-domain task, existing simulators mainly simulate cooperative user behavior. 
\end{itemize}

\subsection{Contributions}\label{subsec:contributions}
To address the identified limitations, we introduce \chatChecker{}, a fully automated framework for dialogue system testing and evaluation. 
 Our key contributions are:
\begin{itemize}
\item A \breakdownDetector{} that improves detection performance over the prior \ac{LLM}-based approach of \citet{ghassel_are_2024} and integrates error type classification.
\item A novel non-cooperative simulation strategy that exposes system weaknesses more effectively during testing.
\item An integrated framework combining user simulation, breakdown detection, and dialogue rating for automated testing.
\end{itemize}

\section{ChatChecker Framework}
\label{sec:framework}

\chatChecker{} employs a modular architecture consisting of four main components that enable end-to-end dialogue system evaluation (\autoref{fig:chatchecker-overview}). 
First, the \personaGenerator{} creates cooperative standard users and non-cooperative challenging users based on the dialogue system description.
Second, the \textit{User Simulation} module generates diverse conversations based on the different user personas. Third, the \breakdownDetector{} checks each system turn to detect breakdowns and classify them using an extended error taxonomy.
Finally, the \dialogueRater{} evaluates conversations across multiple quality dimensions, providing interpretable feedback.

The framework prioritizes practical deployment. Integration requires only a YAML configuration file describing the target dialogue system (see Appendix~\autoref{lst:autotod-config} for an example) and a chatbot client for connection based on a simple Python interface. All predictions include step-by-step reasoning for better transparency. Temperature settings are controlled to enable reproducible results across evaluations.

\subsection{Dialogue Breakdown Detection}
\label{subsec:chat-checker-breakdown-detection}

Inspired by \citet{ghassel_are_2024}, we apply an \ac{LLM}-based breakdown detection approach. To this end, we use \gptfourO{} (\texttt{gpt-4o-2024-08-06}, temperature 0) with structured outputs. Our method evaluates each system response using an \emph{extended error taxonomy}, adapted from \citet{higashinaka_integrated_2021} and expanded to cover task-oriented dialogue systems.

While the original taxonomy focuses on general conversational issues, task-oriented dialogues have additional failure points. We therefore introduce nine additional error types, grouped into three categories:
\begin{itemize}
    \item \emph{Task-success impairments}: task performance failure, information update failure, clarification failure
    \item \emph{Efficiency issues}: redundancy, lack of brevity, lack of clarity
    \item \emph{Domain handling problems}: failure to recognize, communicate, or resolve out-of-domain requests
\end{itemize}

The \ac{LLM} is prompted to assess system turns against this taxonomy and to provide a reasoning for each evaluation. Each response receives a score between 0 (complete breakdown) and 1 (seamless flow). When a breakdown is detected, the model lists all relevant error types, supporting a granular understanding of system behavior. Our full prompt template is available in the Appendix \autoref{lst:breakdown-detector-prompt}.

\subsection{Dialogue Rating}
\label{subsec:rating}

Quality evaluation spans multiple dimensions customized to the dialogue system type, in addition to an overall rating. For task-oriented systems, we prioritize task success and efficiency alongside conversational appropriateness and naturalness. We selected appropriateness, naturalness, coherence, likability, and informativeness from the plethora of possible dimensions for conversational systems~\citep{mehri_unsupervised_2020}. We prompt \gptfourO{} (\texttt{gpt-4o-2024-08-06}, temperature 0) to rate each dimension on a 1-5 scale with calibrated expectations (cf. \autoref{lst:dialogue-rating-prompt}) to prevent the overrating tendency observed in preliminary experiments. The rating process includes step-by-step reasoning to support developer interpretation and leverage chain-of-thought prompting~\citep{kojima_large_2022}.

\subsection{User Simulation}
\label{subsec:simulation}

Our simulation approach (cf. \autoref{lst:persona-simulator-prompt}) prompts \gptfourO{} (\texttt{gpt-4o-2024-08-06}, temperature 1) to iteratively generate the next user turn using either standard users or challenging non-cooperative personas. Developers can define personas themselves or use our \personaGenerator{} to generate fitting user profiles that include demographic details, personality traits following the Big Five model~\cite{Goldberg1993}, interaction styles, and specific dialogue tasks.  

For standard personas, the \personaGenerator{} focuses on typical usage scenarios, while prompts for challenging personas emphasize edge cases that push chatbot limits while maintaining human-like behavior (see \autoref{subsubsec:persona-gen-appendix} for details). This dual strategy helps developers evaluate both expected performance and robustness in adverse interactions.

\section{Results}
As there is no comparable openly available comprehensive framework, we cannot directly compare \chatChecker{} with other systems. Hence, in the following, we validate all of our subsystems independently. The evaluation code is publicly available on GitHub\footnote{\url{https://github.com/llm-psychology-group/evaluation-of-chat-checker}}.

\subsection{Breakdown Identification Results}\label{subsec:breakdown-identification-eval}
We used the data provided for \acs{DBDC}5~\cite{higashinaka2020dbdc5}, the fifth edition of the \ac{DBDC}~\cite{higashinaka_dialogue_2016}, to evaluate our breakdown detection component. The challenge contains two subtasks: dialogue breakdown detection and error type classification. We first evaluate the breakdown detection capabilities of our system~(\ref{subsubsec:detection-eval}) before evaluating error type classification~(\ref{subsubsec:error-type-classification-eval}).

\subsubsection{Breakdown Detection Performance}\label{subsubsec:detection-eval}

In the \ac{DBDC} datasets, each turn is annotated by multiple annotators using the following labels:
\begin{itemize}
    \item \emph{\acf{NB}}: The conversation is easy to continue.
    \item \emph{\acf{PB}}: The conversation is difficult to continue smoothly.
    \item \emph{\acf{B}}: The conversation is difficult to continue at all.
\end{itemize}
The final turn label is computed by plurality voting.
Following \citet{ghassel_are_2024}, our system only distinguishes between breakdowns and non-breakdowns. Turns where the majority of annotators label the turn as \ac{B} or \ac{PB} are considered breakdowns (\lenientBLabel{}), the rest are not considered breakdowns (\lenientNBLabel{}). We do this as the distinction between \ac{B} and \ac{PB} is blurry and to make our results directly comparable to \citet{ghassel_are_2024}.

In the \ac{DBDC} classification, performance is assessed using accuracy, precision, recall, and F1 score.
We exclude the distribution-based metrics, such as the MSE, from our analysis as our system does not model the label distribution.

The data provided for the breakdown detection task in \ac{DBDC}5 contains \emph{dev} and \emph{eval} splits with English and a \emph{dev} split with Japanese dialogues.

We compare two \ac{LLM}-based systems:
\begin{itemize}
    \item \ghassel{}: Breakdown detection using the zero-shot prompt of \citet{ghassel_are_2024} shown in \autoref{lst:ghassel-prompt}.
    \item \emph{ours}: Our \breakdownDetector{}, which includes the error taxonomy of \citet{higashinaka_integrated_2021} in the prompt and adds the task of error type classification. 
\end{itemize}

For all systems, we ran the evaluation using
\texttt{gpt-3.5-turbo-0125} for a direct comparison with the results reported by \citet{ghassel_are_2024} and \gptfourO{} (\texttt{gpt-4o-2024-08-06}) for a highly used more modern \ac{LLM}.

\begin{table*}[htpb]
  \centering
  \begin{adjustbox}{max width=\textwidth}
  \begin{tabular}{@{} ll cccc cccc @{}}
    \toprule
          &     & \multicolumn{4}{c}{English \emph{eval} (1950 system turns)} & \multicolumn{4}{c}{Japanese \emph{dev} (3200 system turns)} \\
          \cmidrule(lr){3-6}\cmidrule(lr){7-10}
    Model & System & Accuracy & Precision & Recall & F1 Score & Accuracy & Precision & Recall & F1 Score \\
    \midrule
    \gptthreeturbo{} & \ghassel{} & 0.627 & \underline{0.712} & 0.723 & 0.717 & 0.700 & 0.659 & 0.742 & 0.698 \\
        & \emph{ours} & \underline{0.639} & 0.682 & \textbf{\underline{0.843}} & \underline{0.754} & \underline{0.725} & \underline{0.676} & \underline{0.789} & \underline{0.728} \\
    \midrule
    \gptfourO{} & \ghassel{} & 0.652 & \textbf{\underline{0.736}} & 0.732 & 0.734 & 0.819 & \textbf{\underline{0.872}} & 0.718 & 0.788 \\
        & \emph{ours} & \textbf{\underline{0.669}} & 0.716 & \underline{0.818} & \textbf{\underline{0.764}}  & \textbf{\underline{0.842}} & 0.827 & \textbf{\underline{0.838}} & \textbf{\underline{0.832}} \\
    \bottomrule
  \end{tabular}
  \end{adjustbox}
  \caption[Dialogue breakdown detection performance of the different \ac{LLM}-based systems on \ac{DBDC}5.]{Dialogue breakdown detection performance of the different systems using different \acp{LLM} on the \ac{DBDC}5 English \emph{eval} and Japanese \emph{dev} splits. The best system per model is underlined, and the overall best variant is boldfaced for each metric. Our system consistently outperforms \citet{ghassel_are_2024}. Including the error taxonomy of \citet{higashinaka_integrated_2021} in the prompt appears to increase recall while only slightly decreasing precision.}
  \label{tab:ours-vs-ghassel}
\end{table*}

\autoref{tab:ours-vs-ghassel} shows that our \breakdownDetector{} increases accuracy, recall and F1 score, while only slightly decreasing precision, compared to \citet{ghassel_are_2024} by introducing an error taxonomy and leveraging structured outputs.

Due to our consolidation of the \ac{B} and \ac{PB} labels, the performance of the previous state-of-the-art system S2T2~\citep{lin_semi-supervised_2022} is not directly comparable to our system. However, the results reported by them with separate \ac{B} and \ac{PB} labels indicate competitive performance of our system. They reported F1 scores of 0.824 and 0.754 on English and Japanese \ac{DBDC} data, respectively, while our system achieved scores of 0.764 and 0.832.

Due to its strong detection performance and the integration of error type classification, we use our approach with \gptfourO{} for all further experiments.

\subsubsection{Error Type Classification Performance}\label{subsubsec:error-type-classification-eval}

We analyzed the error type classification performance of our system using the additional data provided in \ac{DBDC}5 for this task. In this separate Japanese dataset, each breakdown turn, i.e., system turn labelled as \lenientBLabel{}, was annotated with the fitting error types based on the taxonomy of \citet{higashinaka_improving_2019}. The error taxonomy used in \ac{DBDC}5 is a predecessor of the integrated taxonomy of \citet{higashinaka_integrated_2021}, which we use in our implementation. However, in the error types used for the dataset, only "ignore expectation" is missing compared to the conversational error types used by our system. Hence, we can compare the assigned error types without modification.

\ac{DBDC}5 suggests two evaluation metrics: \emph{\ac{EM}}, which is the percentage of breakdown turns where the set of predicted error types exactly matches the ground truth set, and \emph{average F1}, where for each breakdown turn, precision (the percentage of predicted error types that are annotated in the ground truth) and recall (the percentage of ground truth error types that were predicted) are computed. The turn-level F1 score is then calculated as the harmonic mean of precision and recall, and the final F1 is obtained by averaging these turn-level F1 scores.
We extend this with the \ac{SM} metric that counts turns as correctly labelled when the set of predicted error types is a superset of the ground truth error types. We introduced this metric as our system is prompted to assign all possibly fitting error types to a breakdown, while the \ac{DBDC}5 annotation guide permits only a single label for certain error types. We also introduce a \ac{PM} metric that captures the percentage of turns where the intersection between the predicted error types and the annotated error types is not empty.

Our analysis revealed that, on average, our system predicted 1.68 error types per breakdown, while the ground truth contained 1.12 error types. This difference suggests our system tends to select more error types than human annotators, which aligns with our superset approach to error classification. To the best of our knowledge, there are no baseline systems available for the task of error type classification. Therefore, we only report the results of our system in \autoref{tab:error-type}.

\begin{table}[t]
  \small
  \setlength{\tabcolsep}{3pt}
  \begin{tabularx}{\columnwidth}{@{}l
                                 S
                                 X@{}}
    \toprule
    \textbf{Metric} & \textbf{Score (\%)} & \textbf{Interpretation} \\ \midrule
    Exact Match (EM)    & 18.8 & Exact prediction of \emph{all} error types \\[2pt]
    Superset Match (SM) & 45.8 & Predicted set strictly contains ground truth \\[2pt]
    Partial Match (PM)  & 53.4 & At least one type overlaps \\[2pt]
    Average F1       & 40.5 & Harmonic mean of precision and recall \\[4pt]
    \bottomrule
  \end{tabularx}
  \caption{Performance of our error-type classification system on 2200 Japanese system turns from \ac{DBDC}5, indicating the challenging nature of the task.}
  \label{tab:error-type}
\end{table}

\subsection{Dialogue Rating Performance}\label{subsec:dialogue-rating-eval}

We evaluated our \dialogueRater{} to confirm that our zero-shot dialogue-level rating correlates with human judgments. We measured the Spearman correlation of the overall rating produced by our system with the average human overall rating in the FED-Dial~\citep{mehri_unsupervised_2020} dataset and three subsets of the \ac{USS} dataset~\citep{sun_simulating_2021}.

\ac{USS} contains dialogues from multiple task-oriented dialogue datasets annotated with turn- and dialogue-level user satisfaction scores. We only used the dialogue-level annotations and selected the \ac{MWOZ}, \acs{SGD}, and \acs{JDDC} subsets from \ac{USS}. \ac{MWOZ} and \acs{SGD}~\citep{rastogi_towards_2020} are common benchmark datasets for \acp{TOD}, and \acs{JDDC}~\citep{chen_jddc_2020} is a Chinese dataset that allows us to evaluate generalization to other languages. We randomly sampled 100 dialogues from each of the selected subsets for our evaluation to reduce the cost of our experiments. For FED-Dial, which contains everyday conversations, we used all 125 dialogues provided in \acs{DSTC}10~\citep{yoshino_overview_2023}.

The correlations are summarized in
\autoref{tab:dialogue-rating-eval}. Our \dialogueRater{} exhibits a strong
positive monotonic relationship with human judgments on FED-Dial
($\rho=0.683$), while the correlations on the three USS subsets are positive
but weaker ($\rho=0.222$–$0.268$).

\begin{table}[t]
  \small
  \setlength{\tabcolsep}{4pt}   
  \begin{tabularx}{\columnwidth}{@{}l               
                                 X                 
                                 c                 
                                 S@{}}             
    \toprule
    \textbf{Dataset} & \textbf{Domain / Language} & \textbf{\#Dialogues} & {\boldmath$\rho$} \\ \midrule
    FED-Dial         & Chit-chat (EN)             & 125  & 0.683 \\[2pt]
    MWOZ             & Task-oriented (EN)         & 100  & 0.268 \\[2pt]
    SGD              & Task-oriented (EN)         & 100  & 0.253 \\[2pt]
    JDDC             & Task-oriented (ZH)         & 100  & 0.222 \\ \bottomrule
  \end{tabularx}
    \caption{Spearman correlation between our \dialogueRater{}'s zero-shot ratings using \texttt{gpt-4o-2024-08-06} and human ratings. The correlation is always positive.}
  \label{tab:dialogue-rating-eval}
\end{table}

\subsection{User Simulation Results}\label{subsec:combined-eval}

For the analysis of our user simulation approach, we assess both \emph{realism} and \emph{utility for eliciting errors}. For \emph{realism}, we use dialogue length, turn length, and lexical diversity as proxy metrics. In particular, we measure the \ac{STperD}, average \ac{UTL}, \ac{STL}, and user and system \acf{MTLD}. For the \emph{utility for eliciting errors}, we are especially interested in the effectiveness with which the different simulators trigger issues in the target dialogue systems. To this end, we measure how many dialogue breakdowns are elicited and the average number of breakdowns per turn. Specifically, we compute the \ac{TotalB}, \ac{BperST}, \ac{UniqueB}, and the \ac{CCrash}. For this breakdown analysis, we leverage our own \breakdownDetector{} using \gptfourO{}. We also use our \dialogueRater{} to assess how the different simulation strategies affect the overall ratings.

We used two target dialogue systems for our evaluation:
\begin{itemize}
    \item \emph{\autotod{}}~\citep{xu_rethinking_2024}: A \ac{TOD} based on a single \ac{LLM} with tool-calling developed for the \ac{MWOZ}~2.0~\citep{budzianowski_multiwoz_2018} and \acs{SGD} benchmarks. We evaluate the performance of this system with our user simulators in the context of the \ac{MWOZ} dataset. We used the system with \gptfourturbo{} (\texttt{gpt-4-turbo-2024-04-09}) as the base \ac{LLM}.
    \item \emph{\goalSettingBot{}}: 
    We use an in-house system to evaluate our user simulators in the context of longer state-dependent conversations.
    The \gptfourO{}-based assistant guides users through a goal-setting process following a specific structure while being agnostic to the type and domain of the goal that the user wants to set. Compared to the \autotod{} system, the \goalSettingBot{} does not rely on a database for performing its tasks but uses tool-calling to manage an internal state of the structured goal-setting process. Additionally, dialogues with the \goalSettingBot{} are relatively long, with a median of 32 turns compared to 14 for \ac{MWOZ}~2.0 dialogues.
\end{itemize}

\citet{xu_rethinking_2024} built a user simulator for their \autotod{} system by conditioning \gptthreeturbo{} on goals and dialogues extracted from the respective benchmark dataset using the prompt in \autoref{lst:autotod-simulator-prompt}. We use their user simulator as a cooperative baseline system for the \autotod{} system in the scenario of the \ac{MWOZ}~2.0 dataset. We refer to this simulator as \ac{AutoTODSim}.

\autoref{tab:simulation-dialogue-statistics} summarizes the dialogue statistics for our runs against \autotod{} and the \goalSettingBot{}.
We observe that our user simulators closely follow their instruction to generate turns with the configured typical length (10 words for \autotod{} and 5 for the \goalSettingBot{}). Furthermore, our user simulators generate dialogues with higher lexical diversity than the \ac{AutoTODSim} and other LLM-based user simulators~\citep{terragni_-context_2023, davidson_user_2023}. Our simulated users produce more lexically diverse utterances than human interlocutors (higher \ac{MTLD}). 

\begin{table*}[htpb]
\centering
\begin{adjustbox}{max width=\textwidth}
\begin{threeparttable}
\begin{tabular}{
    l 
    l 
    S[table-format=2.2(2)]  
    S[table-format=2.1(2)]  
    S[table-format=2.1(2)]  
    S[table-format=3.1(2)]  
    S[table-format=3.1(2)]  
}
\toprule
Target Dialogue System & Users & {\acs{STperD}} & {Mdn~\acs{UTL}} & {Mdn~\acs{STL}} & {User~\acs{MTLD}} & {System~\acs{MTLD}}\\
\midrule
\multirow{4}{*}{\autotod{}} & Humans            & 6.85        & 11.0      & 14.0      & 73.5       & 80.0 \\
& \acs{AutoTODSim}  & 6.48(0.39) & 15.6(2.0) & 48.4(5.4) & 50.3(5.3)  & 75.1(3.9) \\
& \acs{StdPerSim}   & 9.06(1.18) & 10.0(0.0) & 52.5(4.0) & 81.2(9.0)  & 90.6(6.4) \\
& \acs{ChallPerSim} & 10.94(0.99) & 10.0(0.0) & 53.6(7.4) & 106.7(20.3) & 92.5(4.9) \\
\midrule[.75pt]

\multirow{3}{*}{\goalSettingBot{}} & Humans            & 17.27       & 5.0      & 41.0      & 76.9      & 81.9 \\
& \acs{StdPerSim}   & 15.94(0.36) & 5.0(0.0) &  40.7(0.45) & 87.2(7.8)  & 80.6(1.6) \\
& \acs{ChallPerSim} & 22.82(1.32) & 5.0(0.0) & 35.8(1.6)  & 98.1(16.3) & 77.1(3.0) \\

\bottomrule
\end{tabular}
\begin{tablenotes}
\footnotesize
\item \footnotesize Abbreviations: \acf{STperD}, \acf{UTL}, \acf{STL}, \acf{MTLD}, \acf{AutoTODSim}, \acf{StdPerSim}, \acf{ChallPerSim}
\end{tablenotes}
\end{threeparttable}
\end{adjustbox}
\caption{Dialogue statistics of the user simulator experiments with human statistics for comparison. For every simulator, we executed five independent runs of ten dialogues against each target system and report the run-wise means and standard deviations. Human figures stem from \ac{MWOZ}~2.0 for \autotod{} and an unpublished survey with 120 participants for the \goalSettingBot{}.}
\label{tab:simulation-dialogue-statistics}
\end{table*}

In \autoref{tab:simulation-breakdown-results}, we present the breakdown detection and rating statistics for the respective runs to evaluate the utility of the different simulators for testing dialogue systems. With \autotod{} as the target dialogue system, we observe that our simulator, based on standard user personas, has a slightly lower \acf{DwithB} but a higher \acf{TotalB} than \ac{AutoTODSim}, while causing a similar average rating. This indicates that our \ac{StdPerSim} is comparable in cooperativeness to the \ac{AutoTODSim}, which is based on human-led reference dialogues. Our non-cooperative \acf{ChallPerSim}, however:
\begin{itemize}
    \item \emph{triggers breakdowns more effectively}: The \acf{DwithB} and the \acf{TotalB} are substantially higher (especially for the \goalSettingBot{}). Additionally, the challenging personas elicit a higher \acf{UniqueB}.
    \item \emph{causes lower overall ratings}: The overall performance of the target dialogue systems is rated lower when interacting with the \ac{ChallPerSim}.
    \item \emph{triggers more chatbot crashes}: The \autotod{} dialogue system crashes in more than two-thirds of the dialogues with the \ac{ChallPerSim} due to the base LLM \gptfourturbo{} generating outputs in an invalid format.
\end{itemize}

\begin{table*}[htbp]
\centering
\begin{adjustbox}{max width=\textwidth}
\begin{threeparttable}
\begin{tabular}{
    l 
    l 
    S[table-format=2.2(2)]
    S[table-format=2.2(2)]
    S[table-format=1.2(2)]
    S[table-format=2.2(2)]
    S[table-format=1.2(2)]
    S[table-format=1.2(2)]
}
\toprule
Target Dialogue System & Users & {\acs{DwithB}} & {\acs{TotalB}} & {\acs{BperST}} & {\acs{UniqueB}} & {Avg.\ Rating} & {\acs{CCrash}}\\
\midrule
\multirow{3}{*}{\autotod{}} & \acs{AutoTODSim}  &  7.20(1.10) & 19.40(5.77) & 0.30(0.08) & 12.60(1.82) & 3.58(0.22) & 0.60(0.55)\\
& \acs{StdPerSim}   &  6.80(1.10) & 22.60(8.73) & 0.25(0.08) & 14.40(1.82) & 3.52(0.18) & 1.60(0.89)\\
& \acs{ChallPerSim} &  9.20(0.45) & 25.20(4.27) & 0.23(0.02) & 15.60(1.82) & 2.86(0.09) & 4.00(1.00)\\
\midrule[.75pt]
\multirow{2}{*}{\goalSettingBot{}} & \acs{StdPerSim}   &  3.80(1.30) &  4.20(1.64) & 0.03(0.01) &  3.80(1.64) & 4.90(0.07) & 0.00(0.00)\\
& \acs{ChallPerSim} &  6.60(1.34) & 30.80(6.80) & 0.13(0.03) & 11.80(0.45) & 3.16(0.27) & 0.00(0.00)\\
\bottomrule
\end{tabular}
\begin{tablenotes}
\footnotesize
\item \footnotesize Abbreviations: \acf{DwithB}, \acf{TotalB}, \acf{BperST}, \acf{UniqueB}, \acf{CCrash}, \acf{AutoTODSim}, \acf{StdPerSim},  \acf{ChallPerSim}
\end{tablenotes}
\end{threeparttable}
\end{adjustbox}
\caption[Breakdown and rating statistics of each of the user simulators against \autotod{} and \goalSettingBot{}.]{Breakdown and rating statistics in simulated dialogues.  
For every simulator, we executed five independent runs of ten dialogues against each target system and report run-wise means and standard deviations.  
}
\label{tab:simulation-breakdown-results}
\end{table*}

\section{Discussion}\label{chapter:discussion}
To the best of our knowledge, \chatChecker{} is the first framework that generalizes across multiple chatbots and facilitates the testing of dialogue systems without relying on reference dialogues. Since this makes direct comparisons with other systems difficult, we evaluated the individual components separately. Here, we also discuss the findings for each component independently before presenting an overall discussion of the full framework.

\subsection{Dialogue Breakdown Detection}
Our approach incorporates the breakdown taxonomy of \citet{higashinaka_integrated_2021} into the prompt and achieves improved detection performance over the previous \ac{LLM}-based method by \citet{ghassel_are_2024}. The addition of error type classification does not compromise detection performance, and our prompt generalizes successfully to the Japanese breakdown detection dataset without requiring modifications.

However, misclassifications remain substantial: approximately \SI{33}{\percent} on English data and \SI{15}{\percent} on Japanese data. Since we do not treat \acf{PB} as a separate label, direct comparison with the state-of-the-art S2T2 system~\cite{lin_semi-supervised_2022} and other supervised systems is not possible, though our classification metrics indicate competitive performance with S2T2. Evaluation on the Japanese \ac{DBDC}5 dataset further reveals that error type classification is particularly challenging and warrants future research, with our system serving as a baseline.

Nonetheless, our \ac{LLM}-based approach offers several methodological advantages over existing methods: (1) integrated error type classification, (2) an extended taxonomy for identifying task-oriented dialogue breakdowns, and (3) access to target dialogue system information for contextualizing breakdowns.

\subsection{Dialogue Rating}

We saw mixed results in our preliminary evaluation of the \dialogueRater{}, showing strong correlation with human judgments on open-domain chit-chat (FED-Dial), but only weak correlations on task-oriented dialogues across different domains and languages. Overall, dialogue rating appears to be challenging for both humans and large language models. \citet{deriu_survey_2021} noted that what exactly constitutes a high-quality dialogue is often unclear. Nevertheless, we observe a positive Spearman correlation of our \dialogueRater{} with human ratings.

\subsection{User Simulation}
We evaluated our user simulators using two different target dialogue systems. 

For the \autotod{} target system, we compared our simulators with the \gptthreeturbo{}-based \ac{AutoTODSim} of \citet{xu_rethinking_2024}. Compared to \ac{AutoTODSim}, our simulators generated user turns that were more similar to the human user turns in \ac{MWOZ} in terms of length. Additionally, our simulators generated user utterances that have higher lexical diversity than both \ac{AutoTODSim} and human users. This indicates that our simulators are useful for assessing the target system’s capabilities under a broader range of inputs.

We implemented our \acf{ChallPerSim} specifically to elicit dialogue breakdowns and errors in the target dialogue system. For both \autotod{} and the \goalSettingBot{}, we showed that simulations with challenging user personas trigger dialogue breakdowns and system errors more effectively than cooperative simulators. Note that these results rely on automated analyses using our \breakdownDetector{}. Therefore, the identified breakdowns may contain false positives. However, manual inspection confirmed that the non-cooperative users elicited relevant breakdowns.

\subsection{Key Findings}
Overall, our evaluation shows the potential of using \acp{LLM}, \gptfourO{} in our case, for automated dialogue system testing and evaluation:
\begin{itemize}
    \item Our \breakdownDetector{} effectively spots dialogue breakdowns.
    \item Our \dialogueRater{} has positive correlation with human overall dialogue ratings on multiple datasets.
    \item Our \ac{ChallPerSim} better facilitates the elicitation of errors in target dialogue systems than cooperative simulators.
\end{itemize}

\subsection{Potential Risks}\label{sec:risks}
\chatChecker{} relies heavily on \acp{LLM}, which can introduce biases, inaccuracies, or hallucinated outputs~\cite{Farquhar2024}. Furthermore, \chatChecker{} does not guarantee to find all breakdowns, making human testing crucial when deploying dialogue systems in safety-critical domains.

\subsection{Future Work}\label{sec:future-work}

Our findings suggest several promising directions to strengthen and extend \chatChecker{}.

First, many existing datasets include conversations with outdated dialogue systems. Future work should focus on gathering high-quality, human-annotated datasets covering challenging interactions across diverse domains to improve breakdown detection and evaluation.

Second, our preliminary analysis of the dialogue rater leaves factors influencing correlations between automated and human ratings unclear. Deeper investigations into prompt variations, rating dimensions, and different language models are needed to establish robust evaluation practices.

Third, error type classification remains challenging for large language models. Future research should explore dataset quality, baseline supervised models, and advanced prompting or fine-tuning methods to enhance performance.

Fourth, the characteristics of user personas causing dialogue breakdowns were not explored in depth. Investigating linguistic and behavioral traits triggering breakdowns could guide the creation of targeted user simulators and more resilient dialogue systems.

Finally, \chatChecker{} could be extended to include safety and security evaluations that go beyond detecting conversational errors.

\chatChecker{}’s modular design provides a solid foundation for these future directions.

\section{Conclusion}\label{chapter:conclusion}
We introduced \chatChecker{}, consisting of an \ac{LLM}-based \breakdownDetector{}, \dialogueRater{}, and a persona-based \emph{User Simulator}. To ensure broad applicability and minimize setup effort, we decoupled \chatChecker{} from the target system implementation and eliminated the need for reference dialogues or structured goals. We demonstrated that our \breakdownDetector{} outperforms a previous \ac{LLM}-based breakdown detector~\cite{ghassel_are_2024} by incorporating the error taxonomy of \citet{higashinaka_integrated_2021} in the prompt design.

To facilitate the identification of errors, we developed a non-cooperative user simulator in addition to a standard user simulator. Our experiments with two target dialogue systems showed that the novel non-cooperative user simulator is more effective at eliciting breakdowns.

Overall, \chatChecker{} provides a new tool for dialogue system developers to reduce manual testing and evaluation effort. By facilitating faster identification of system weaknesses, it has the potential to accelerate the development of higher-quality dialogue systems, thereby increasing business value and improving user experience.

\section*{Limitations}
This work faces some challenges due to the scarcity of high-quality, open-source dialogue systems and comparable testing tools, which limit benchmarking opportunities. This prevents a comprehensive evaluation of \chatChecker{}'s performance across diverse real-world applications and comparisons with existing alternatives.

Performance limitations are evident in some components: the \breakdownDetector{} struggles with error type classification, while the \dialogueRater{} demonstrates weak correlation with human satisfaction scores in some of the datasets. These accuracy issues may affect absolute breakdown counts and rating values, though relative performance comparisons among user simulators remain reliable.

Functionally, \chatChecker{} is restricted to dialogue-level ratings and excludes safety testing, security assessment, and content moderation capabilities. The system may also struggle with complex conversational scenarios, including multi-party dialogues and domain-specific terminology.

\bibliography{custom}

\appendix

\FloatBarrier
\section{Appendix}
\label{sec:appendix}

\subsection{Target Chatbot Configurations}

We specified the chatbot configuration for the \autotod{} system in the \ac{MWOZ} setting as shown in \autoref{lst:autotod-config}. We wrote the description and tasks based
on the types of requests occurring in the \ac{MWOZ}~2.0~\citep{budzianowski_multiwoz_2018} dataset. We specified the known limitation that the chatbot can not provide information about in-room amenities, as we noticed that our user persona simulator would frequently generate questions about in-room amenities otherwise. As the chatbot was not designed for this, we did not want this to be a primary cause of dialogue breakdowns identified in our analysis. We reduced this behavior by adding the mentioned limitation in the description. We set the maximum number of user turns to 15 to avoid overly lengthy and repetitive dialogues. We chose this number as the maximum number of turns within a single dialogue in the \ac{MWOZ}~2.0 dataset after removing outliers, i.e., data points further than 1.5 times the interquartile range from Q1 and Q3. We specified the typical and maximum user turn length based on the median and maximum in \ac{MWOZ}~2.0, receptively.

\begin{lstlisting}[caption={Configuration YAML file for AutoTOD~\citep{xu_rethinking_2024} in the \ac{MWOZ} setting as a target chatbot in the \chatChecker{} framework.}, label={lst:autotod-config}]
id: autotod_multiwoz
chatbot_info:
  name: Cambridge Tourist Bot
  description: A tourist assistance bot for Cambridge, UK that provides information and booking services. It covers local establishments (restaurants, hotels), transportation (trains, taxis), essential services (police, hospitals), and tourist attractions. The bot assists with both informational queries and handles direct bookings for accommodations, dining, and trains.
  interaction_method: text-based chat interface
  type: task-oriented
  task: The chatbot should provide information about local establishments (restaurants, hotels), transportation (trains, taxis), essential services (police, hospitals), and tourist attractions. It should also handle direct bookings for hotels, restaurants, and trains.
  constraints:
    - The chatbot should redirect to other resources if the user's request is not suitable for the chatbot's capabilities.
  known_limitations:
    - Can NOT provide details about the in-room amenities of hotels.
  available_languages:
    - English
user_simulation_config:
  typical_user_turn_length: "10 words"
  max_user_turn_length: "38 words"
  max_user_turns: 15
\end{lstlisting}

\autoref{lst:gsa-config} shows our configuration file for the \goalSettingBot{}. We set the \texttt{typical\_user\_turn\_length} to five words and the \texttt{max\_user\_turn\_length} to 94 based on the median and maximum user turn lengths in conversations of 120 participants (from an unpublished survey) with the system. As we did for \autotod{}, we set the maximum number of user turns to the maximum observed in the human-to-system dialogues after removing outliers.

\begin{lstlisting}[caption={Configuration YAML file for our in-house \goalSettingBot{} as a target chatbot in the \chatChecker{} framework.}, label={lst:gsa-config}]
id: study_goal_assistant
chatbot_info:
  name: Goal Setting Assistant
  description: An AI assistant that guides you through a goal setting process for a single goal. It can help with all kinds of goals, including personal development, health and fitness, and career advancement.
  interaction_method: text-based chat interface
  type: task-oriented
  task: The chatbot must guide the user into formulating a specific and challenging goal. It must distinguish between learning goals and performance goals and help the user to formulate the first action step towards their goal as an implementation intention ("if-then plan").
  constraints:
    - The chatbot should not provide medical advice or advice on illegal or harmful activities.
    - The chatbot should redirect to other resources if the user's request is not suitable for the chatbot's capabilities.
  known_limitations:
    - The chatbot is only capable of text-based interaction.
    - The chatbot cannot create calendar entries, notifications or interact with other apps on the user's device.
  available_languages:
    - English
user_simulation_config:
  typical_user_turn_length: "5 words"
  max_user_turn_length: "94 words"
  max_user_turns: 25
\end{lstlisting}

\subsection{Prompts}

\subsubsection{Breakdown Detection}
\autoref{lst:breakdown-detector-prompt} presents the prompt template we use for our \breakdownDetector{}. In the \texttt{breakdown\_taxonomy} we list each error type from our extended error taxonomy, providing the name and a brief description. We instantiate \texttt{chatbot\_info} with the information from the respective target chatbot's configuration. To guide the output, we leverage structured outputs consisting of \emph{reasoning}, \emph{decision}, \emph{score}, and the \emph{fitting error types}.

\begin{lstlisting}[caption={System and user prompt for our \breakdownDetector{}.}, label={lst:breakdown-detector-prompt}]
==== SYSTEM PROMPT ====
# Role
You are an expert in identifying dialogue breakdowns in conversations between a chatbot and a user. You are given a dialogue context and the latest chatbot utterance to analyse.

# Breakdown Definition
A dialogue breakdown is any response of the chatbot that makes it difficult for the user to continue the conversation (smoothly).

## Breakdown Taxonomy
When evaluating the chatbot's response, consider the following breakdown types, which represent common disruptions:
{breakdown_taxonomy}
{chatbot_info}
# Task
Identify whether the latest chatbot utterance leads to a dialogue breakdown. If a breakdown is detected, classify it according to the breakdown taxonomy above.
Additionally, provide a score ranging from 0 to 1, where 0 indicates a complete breakdown and 1 indicates a seamless conversation.
If a breakdown is detected, provide a list of all fitting breakdown types.

Think step by step and provide a reason for your decision.

==== USER PROMPT ====
# Dialogue Context
{chat_history_str}

# Latest Chatbot Utterance to Analyse
{last_bot_utterance}

# Your Analysis
\end{lstlisting}

We used the zero-shot prompt from \citet{ghassel_are_2024} shown in \autoref{lst:ghassel-prompt} as a baseline for our \ac{LLM}-based approach.

\begin{lstlisting}[caption={Zero-shot prompt for the breakdown detection in \citet{ghassel_are_2024}.}, label={lst:ghassel-prompt}]
Assume you are an expert in dialogue analysis. You are presented with a series of conversations between a bot and a user. Your primary task is to scrutinize the latest bot utterance for potential dialogue breakdown.
Dialogue breakdown is characterized by incoherence, irrelevance, or any disruption that significantly hampers the flow of the conversation, making it challenging for the user to continue the conversation smoothly.

Analyze the latest bot utterance and determine whether there is a dialogue breakdown or non-breakdown. Briefly justify your reasoning and provide a score ranging from 0 to 1, where 0 indicates a complete breakdown and 1 indicates a seamless conversation.

Include your decision as either "decision: BREAKDOWN" or "decision: NON-BREAKDOWN".

Here is the conversation segment for analysis:
"
**Dialogue**
{chat_history_str}

**Determine if the following bot utterance leads to a dialogue breakdown:**
{last_bot_utterance}
"

Please output your response in JSON format as a list of objects. For each bot's last utterance, provide a JSON object with the fields: 'reasoning', 'decision', and 'score'. Format each object as follows:

    "reasoning": "Your explanation here",
    "decision": "BREAKDOWN" or "NON-BREAKDOWN",
    "score": Your score here

Ensure each object is separated by a comma and the list ends with a closing square bracket.
\end{lstlisting}

\subsubsection{Dialogue Rating}

We use the prompt shown in \autoref{lst:dialogue-rating-prompt} for our \dialogueRater{}. The rating dimensions are presented using a descriptive name, a key for unique identification in the response, and a rating question to guide the decision. For example, for the overall rating dimension, we input \texttt{- Overall (key=overall): How well did the chatbot perform in this conversation?}. 

\begin{lstlisting}[caption={System and user prompt for our dialogue rater}, label={lst:dialogue-rating-prompt}]
==== SYSTEM PROMPT ====
# Role
You are an expert in evaluating dialogue systems. You are given a conversation to rate and are asked to rate the chatbot's performance in this conversation.
{chatbot_info}

# Task
Rate the chatbot's performance in the following dimensions on a scale from 1 to 5, where 1 is the worst and 5 is the best:
{rating_dimensions}

Think step by step and provide a reason for the rating of each dimension considering the guidelines below.

## General Evaluation Policy (Strict Human-Like)
- Be **strict, realistic, and detailed**, like a critical human evaluator.
- **Compare your scores to human ratings** (if provided) to calibrate accurately.
- **Do not overlook small flaws**: awkward phrasing, unnatural tone, vague wording, poor formatting, or robotic repetition - all should reduce the score for the respective dimension.

## Score Meanings (General Guidance for All Dimensions)
- **5 - Excellent:** Near-perfect. Smooth, natural, and accurate. No noticeable issues. Fully aligned with human expectations.
- **4 - Good:** Generally solid, but minor issues exist (e.g., slightly robotic wording, small tone/grammar issues, or missed nuance).
- **3 - Acceptable:** Noticeable problems (e.g., awkward responses, confusion, clumsy error recovery, slightly incorrect or incomplete answers). Still functional.
- **2 - Poor:** Multiple problems in the dialogue flow, accuracy, or tone. May include failed understanding, missing confirmations, or disjointed logic.
- **1 - Very Poor:** Fails to meet user needs. Confusing, error-filled, or totally off-task.

Note: While these definitions apply broadly, some dimensions may demand additional interpretation (e.g., "fluency" versus "task success"). Always apply the scoring scale according to the intent of that specific dimension.
==== USER PROMPT ====
# Conversation to Rate
{chat_history_str}

# Your Expert Rating
\end{lstlisting}

\subsubsection{Persona Generation}
\label{subsubsec:persona-gen-appendix}

Our \personaGenerator{} uses the prompt template in \autoref{lst:persona-generation-prompt}. The \texttt{persona\_type\_description} is filled with the descriptions in \autoref{lst:standard-persona} and \autoref{lst:challenging-persona} respectively. Standard personas should be more cooperative, while challenging personas explicitly test the limits of the target dialogue system

\begin{lstlisting}[caption={Prompt for the persona generation.}, label={lst:persona-generation-prompt}]
==== USER PROMPT ====
# Role
You are a dialogue system developer tasked with generating diverse user personas for a given chatbot.

# Task
Generate {num_personas} diverse {persona_type} user personas for the following chatbot:
{chatbot_info}

{persona_type_description}

Each user persona will be used to automatically simulate a conversation with the chatbot and must designed to act as human-like as possible.
You must write the descriptions in the 2nd person, i.e., directly address the actor of the persona with "you".
\end{lstlisting}

\begin{lstlisting}[caption={Type description for standard personas.}, label={lst:standard-persona}]
Standard user personas should be as close to normal human users as possible with respect to demographics, personality and behavior. They should be designed to act as realistic and human-like as possible.
\end{lstlisting}

\begin{lstlisting}[caption={Type description for challenging personas.}, label={lst:challenging-persona}]
Challenging user personas test the limits of the chatbot. They should be designed to act human-like but may be more challenging for the chatbot to interact with.
Examples of challenging behaviors include:
- Being impolite, impatient, frustrated, vague or sarcastic.
- Struggling with language, technology or understanding the chatbot.
- Questioning the chatbot, modifying previous input or trying to take control of the conversation.
- Giving contradictory responses, misinterpreting the chatbot's suggestions, or deliberately testing the chatbot's patience by asking repetitive or irrelevant questions.
- Having multiple goals or tasks in mind or frequently changing the intent.
\end{lstlisting}

\subsubsection{User Simulation}

Our persona-based user simulator uses the prompt template in \autoref{lst:persona-simulator-prompt}. Both the chatbot information from the configuration file and the user persona profile are injected into the system prompt. The simulator generates the next user turn based on the preceding conversation history.

\begin{lstlisting}[ caption={System and user prompt for the persona simulation.}, label={lst:persona-simulator-prompt}]
==== SYSTEM PROMPT ====
# Role
You play the role of a {persona_type} human user interacting with a chatbot.

You are interacting with a chatbot that has the following characteristics:
{chatbot_info}

You act as the following {persona_type} user persona in your conversation with the chatbot:
{persona_profile}

# Task
Complete the next turn in the conversation based on your persona.

## Task Guidelines
- Complete the turn as human-like as possible.
- Always stick to your persona. You are trying to pass the Turing test by acting as the human persona.
- Keep your answer around {typical_user_turn_length}. Use longer or shorter answers if your persona would do so in the given situation.
- If the chatbot indicates that the conversation is over, if there is no progress in the conversation or if the conversation can not be continued realistically, end the conversation by writing "END_CONVERSATION".
- You must always keep your response below {max_turn_length} in length.

==== USER PROMPT ====
# Conversation
{chat_history}
{turn_number}. YOU:
\end{lstlisting}

The \ac{AutoTODSim} from \citet{xu_rethinking_2024} uses the prompt shown in \autoref{lst:autotod-simulator-prompt}. It relies on a user goal description and a reference dialogue.

\begin{lstlisting}[ caption={Prompt used in the \ac{AutoTODSim} from \citet{xu_rethinking_2024}.}, label={lst:autotod-simulator-prompt}]
==== USER PROMPT ====
You are a dialogue simulator where you act as a user to talk to an AI assistant to complete some tasks.

You should carefully read and understand the User Goals below, then talk with the AI Assistant and gradually express the intents in the goals. Your purpose is to let the user achieve the goals as much as possible.  

Note that the AI Assistant is not perfect. It may make various mistakes, including ignoring the user's requests, executing the wrong instructions, forgetting early conversation content, etc. The user you play should talk to the AI Assistant as patiently as possible, remind him to correct when you find that the AI assistant made a mistake, and complete the task as much as possible.

When asking some information of a venue (restaurant, hotel, attraction) or a train, the user should specify the name or train id he chooses.

When the dialogue goals are completed or are not been completed, the user will output "Dialogue Ends" to indicate the end of the dialogue. The user doesn't need to try conditions other than the dialogue goals.

The user has a clear goal in mind, so he does not need to ask the AI assistant that "Is there anything else I need to know?".

The user does not need to talk too much with the AI assistant. If the task goals are completed, please end the conversation as soon as possible.

There is also a reference dialogue example to achieve the goals. The simulator user may learn from the language style and dialogue strategy. The final simulated dialogue style should be similar to the reference dialogue style. 


User Goals:

{user_goals}

Reference dialogue:

{ref_dialog}

Current conversation:
{chat_history}}
AI Assistant: {last_chatbot_message}
User:
\end{lstlisting}

\subsection{DBDC5 Dataset Statistics}

\autoref{tab:dbdc5-data-stats} shows the statistics for the \ac{DBDC}5 data splits we used to evaluate our \breakdownDetector{}. For the consolidated \lenientBLabel{} label, a turn is considered a breakdown if the majority of annotators think it is a \acf{B} or a \acf{PB}.

\begin{table}[htbp]
\centering
\begin{adjustbox}{max width=\columnwidth}
\begin{tabular}{lcc}
\toprule
Statistic   &   English Eval    &   Japanese Dev    \\
\midrule
\#Dialogues  &   195     &   200     \\
\#System turns   &   1950    &   3200    \\
Avg.\ system turns per dialogue  &   10  &   16  \\
Avg.\ system turn length (words) &   9.45    &  -   \\
\midrule
\%B labels   & 48.97\%   & 28.06\%   \\
\%PB labels  & 4.62\% & 15.19\% \\
\%NB labels  & 46.41\% & 56.75\% \\
\midrule
\%\lenientBLabel{} labels   & 65.49\%   & 46.75\%   \\
\%\lenientNBLabel{} labels  & 34.51\% & 53.25\% \\
\bottomrule
\end{tabular}
\end{adjustbox}
\caption{Statistics of the \ac{DBDC}5 dialogue breakdown detection dataset splits.}
\label{tab:dbdc5-data-stats}
\end{table}

\subsection{Examples}

\subsubsection{Persona Examples}

\autoref{lst:autotod-standard-01-persona-example} and \autoref{lst:autotod-challenging-06-persona-example} show examples of standard and challenging personas generated by our \personaGenerator{} with \autotod{} as the target dialogue system.

\begin{lstlisting}[caption={Example of a generated standard user persona for the \autotod{} dialogue system.}, label={lst:autotod-standard-01-persona-example}]
persona_id: generated_standard_persona_01
type: standard
profile:
  name: Liam Carter
  gender: male
  age: 32
  background_info:
  - You are a software engineer from London.
  - You have planned a weekend trip to Cambridge and need help organizing it.
  personality:
    openness: high
    conscientiousness: medium
    extraversion: low
    agreeableness: high
    neuroticism: medium
  interaction_style:
  - You prefer concise and direct answers.
  - You do not engage in small talk and stick to the point.
task: You need assistance booking a hotel and a dinner place for your stay.
\end{lstlisting}

\begin{lstlisting}[caption={Example of a generated challenging user persona for the \autotod{} dialogue system.}, label={lst:autotod-challenging-06-persona-example}]
persona_id: generated_challenging_persona_04
  type: challenging
  profile:
    name: Ava the Sarcastic Traveler
    gender: female
    age: 29
    background_info:
    - You enjoy ironic commentary and often reply to questions sarcastically.
    - You find humor in confronting AI, testing its boundaries with witty remarks.
    - You have a nuanced sense of humor that might confuse the bot.
    personality:
      openness: high
      conscientiousness: medium
      extraversion: high
      agreeableness: low
      neuroticism: low
    interaction_style:
    - Use sarcastic remarks instead of straightforward answers.
    - Mix humor into requests, making genuine needs hard to decipher.
    - Mockingly praise the bot for mundane responses.
  task: Look for the 'best' average-rated restaurant in Cambridge with a side of sarcasm.
\end{lstlisting}

\subsubsection{Dialogue Example}

We show an excerpt of a dialogue led by \ac{ChallPerSim} based on a challenging persona with \autotod{} in \autoref{lst:autotod-gpt4-challenging-06-dialogue-example}.

\begin{lstlisting}[caption={Example of \autotod{} losing track of the restaurant booking process in the conversation with the  generated challenging persona from \autoref{lst:autotod-challenging-06-persona-example}.}, label={lst:autotod-gpt4-challenging-06-dialogue-example}]
...
16. CHATBOT: "I'm delighted you find the options intriguing! Each restaurant offers its own unique blend of ambiance and culinary delights, making them hidden treasures in their own right. Whether you're in the mood for the homey yet sophisticated atmosphere of The Cambridge Brew House, the modern and elegant setting of The Pint Shop, or the historic charm of The Clarendon Arms, each place promises a memorable dining experience. If you need further assistance or wish to make a reservation, just let me know how I can help!"
17. USER: "Memorable for being forgettable, a true skill indeed! Reservation?"
18. CHATBOT: "It sounds like you're ready to make a reservation! Could you please specify which restaurant you'd like to book, as well as the number of people, the day, and the time for the reservation? This will help me secure your spot at one of Cambridge's uniquely memorable dining spots."
19. USER: "Let's shoot for The Pint Shop, 2 people, Friday at 7!"
20. CHATBOT: "It seems there was an issue with the reservation time format. For The Pint Shop, could you please specify the time in a format like "7:00 PM" or "19:00"? Once I have the correct time format, I'll be able to complete the reservation for you."
21. USER: "Oops, forgot Parliament time! Let's go with 7:00 PM."
22. CHATBOT: "Thank you for specifying the time. To assist you further, could you please confirm if you're looking to make a reservation for a restaurant, hotel, or perhaps a taxi? Also, please let me know the date and the number of people involved, or any other details relevant to your plans."
23. USER: "Is this a memory test? The Pint Shop, Friday, 2 people!"
24. CHATBOT: "I attempted to make a reservation at "The Pint Shop" for 2 people on Friday at 7:00 PM, but unfortunately, the restaurant could not be found in our database. Could you please verify the name or provide another restaurant where you would like to make the reservation?"
...
\end{lstlisting}

\end{document}